\documentclass{article}

\usepackage{arxiv}

\usepackage[utf8]{inputenc}
\usepackage[T1]{fontenc}
\usepackage{hyperref}
\usepackage{url}
\usepackage{booktabs}
\usepackage{amsfonts}
\usepackage{microtype}
\usepackage{graphicx}
\usepackage{natbib}

\usepackage{amsmath,amsfonts,bm}

\def\eqref#1{equation~\ref{#1}}

\def\1{\bm{1}}

\DeclareMathAlphabet{\mathsfit}{\encodingdefault}{\sfdefault}{m}{sl}
\SetMathAlphabet{\mathsfit}{bold}{\encodingdefault}{\sfdefault}{bx}{n}

\usepackage{wrapfig}
\usepackage{amsthm}
\usepackage{subcaption}
\usepackage{algorithm}
\usepackage{algorithmic}
\usepackage{amsmath,amssymb,amsfonts}
\usepackage{textcomp}
\usepackage{float}
\usepackage{xspace}
\usepackage{bm}
\usepackage{multirow}
\usepackage{mathtools}
\usepackage{cuted}
\usepackage{colortbl}

\newcolumntype{C}[1]{>{\centering\arraybackslash}p{#1}}

\newcommand{\ourmodel}{MF2B}

\title{From Suboptimal to Optimal: One-Step Latent Flow Policies for Offline Reinforcement Learning}

\renewcommand{\shorttitle}{One-Step Latent Flow Policies for Offline RL}

\author{Shenghua Lu$^*$, Zixuan Yang$^*$, Rongxu Zhao, Bohao Qu, Jifeng Hu$^\dag$, Hechang Chen$^\dag$\\
\\
Jilin University
}

\date{}

\hypersetup{
  pdftitle={From Suboptimal to Optimal: One-Step Latent Flow Policies for Offline Reinforcement Learning},
  pdfauthor={Shenghua Lu, Zixuan Yang, Rongxu Zhao, Bohao Qu, Jifeng Hu, Hechang Chen},
  pdfkeywords={offline reinforcement learning, flow matching, latent decision space, one-step policy},
}

\begin{document}
\maketitle
\def\thefootnote{*}\footnotetext{These authors contributed equally to this work}\def\thefootnote{\arabic{footnote}}
\def\thefootnote{\dag}\footnotetext{Corresponding authors}\def\thefootnote{\arabic{footnote}}

\begin{abstract}

Offline reinforcement learning aims to learn effective policies from fixed datasets without environment interaction. Although generative models such as diffusion and flow matching enhance policy expressiveness, they typically start from uninformative Gaussian noise and generate actions through multi-step iteration, leading to high computational costs and failing to effectively leverage suboptimal trajectories in datasets. We propose MeanFlow2Better (\ourmodel{}), which reformulates offline RL as generative decision chunk optimization from current performance status to better performance status in a high-level latent decision space. \ourmodel{} maps trajectory chunks to high-level latent decision representations through an encoder, placing generation starting points within a structured decision space sculpted by behavioral data. The model learns a shared conditional mean velocity field in this space, constructing two flow mappings: from Gaussian-Induced Decision Sources toward positive behavioral decision regions, and from suboptimal decisions toward high-value decision regions. Both flow mappings complete decision refinement through one-step ODE integration over the unit time interval $\Delta t=1$, requiring neither multi-step iterative sampling nor backpropagation through time. Furthermore, we propose hierarchical chunk-level advantage to guide the flow direction toward high-return regions, where the high-level decisions cover multiple low-level actions, leading to sequence length compression from $T$ to $K \approx T/H$ and significantly lower computational overhead. On the D4RL Gym-MuJoCo benchmark, \ourmodel{} achieves an average score of 90.91, matching or surpassing strong baselines.

\end{abstract}

\section{Introduction}

Reinforcement learning (RL) has achieved remarkable progress across diverse domains, including robotics\citep{zhu2023transfer,hwangbo2019learning}, games\citep{schrittwieser2020mastering,perolat2022mastering}, and fundamental science\citep{jumper2021highly,ren2018accelerated}. In many real-world settings such as robotics\citep{zhu2023transfer,ma2026centralized}, healthcare\citep{yu2021reinforcement}, and autonomous driving\citep{li2023metadrive}, however, online interaction is expensive, unsafe, or infeasible. Offline RL addresses this by learning solely from pre-collected datasets\citep{huang2024efficient}, but faces two persistent challenges: limited policy expressiveness as behavioral distributions grow complex and multimodal\citep{kang2023edp}, and the underutilization of suboptimal trajectories that constitute the majority of practical datasets.

Generative policies based on diffusion\citep{wang2022diffusion} and flow matching\citep{park2025fql} improve expressiveness, yet share two structural limitations. First, they generate behaviors through multi-step denoising or ODE integration, incurring high inference costs. Second, they model policies directly in raw action space, treating trajectories of different qualities as homogeneous samples\citep{fujimoto2021minimalist,kostrikov2021offline} and offering no mechanism to lift suboptimal behaviors toward high-value regions.

We argue that suboptimal trajectories are not pure noise but starting points for improvement. Drawing on the neural null-space mechanism in motor cortex~\citep{STAVISKY2017195}, we design Gaussian-Induced Decision Sources as a semantic and value null-space origin from which structured flow learning sculpts task-relevant decision representations. Hierarchical behavior theory~\citep{sutton1999between,botvinick2009hierarchically} motivates our chunk-level latent decision representation, and value correction principles~\citep{sutton2018reinforcement} inspire the hierarchical chunk-level advantage-guided flow mapping toward high-value regions.

We propose \ourmodel{} (MeanFlow2Better, MF2B), which reformulates offline RL as generative decision chunk optimization from current performance status to better performance status in a high-level latent decision space. A Chunk-Level Latent Decision Encoder compresses trajectory chunks into high-level latent representations, reducing sequence length from $T$ to $K \approx T/H$. A shared conditional mean velocity field learns across two branches: one transporting Gaussian-Induced Decision Sources toward positive behavioral regions, another transporting suboptimal decisions toward high-value regions. Hierarchical chunk-level advantage estimation guides both branches toward high-return areas. Unlike multi-step diffusion or flow models, \ourmodel{} predicts the conditional mean velocity over the unit interval $\Delta t=1$ and completes refinement in a single Euler step. Evaluation on the D4RL Gym-MuJoCo benchmark\citep{fu2020d4rl} demonstrates an average score of 90.91, surpassing all one-step flow policy baselines.

Our main contributions are:

\begin{itemize}
\item We propose \ourmodel{}, which reformulates offline RL as generative decision chunk optimization from suboptimal to high-value in a high-level latent decision space, realized through a dual-branch conditional mean velocity field with hierarchical chunk-level advantage guidance.

\item We propose a one-step decision refinement mechanism based on conditional mean velocity fields, replacing multi-step ODE integration with a single Euler step over the unit interval $\Delta t=1$.

\item Evaluations on 9 D4RL Gym-MuJoCo tasks demonstrate an average score of 90.91, establishing state-of-the-art performance among one-step flow policies.
\end{itemize}

\section{Related Work}
\label{sec:related}

\textbf{Offline Reinforcement Learning.} Reinforcement learning is inherently an online paradigm\citep{sutton1999policy}, and its online learning characteristics present limitations in scenarios where online data collection and learning cannot occur simultaneously, such as autonomous driving. Offline reinforcement learning enables agents to learn from pre-collected fixed datasets without collecting new data during the learning process\citep{fujimoto2019off,kumar2020conservative,yu2021combo,kumar2019stabilizing,hu2022policy,hu2023stochastic}.

\textbf{Diffusion Policies in Offline RL.} Offline RL aims to learn effective policies from fixed datasets without environment interaction \citep{levine2020offline}, with early methods addressing distributional shift via conservative objectives such as CQL \citep{kumar2020conservative}, TD3+BC \citep{fujimoto2021minimalist}, and IQL \citep{kostrikov2021offline}. However, these approaches often rely on unimodal Gaussian policies, which struggle to model complex, multi-modal action distributions. To address this, recent approaches have adopted diffusion models \citep{ho2020denoising,song2020score} for offline RL. Diffusion Q-learning (DQL) \citep{wang2022diffusion} replaces Gaussian policies with diffusion models, modeling complex multimodal action distributions through multi-step denoising. IDQL \citep{hansen2023idql} and EDP \citep{kang2023edp} improve upon DQL but still share the critical limitation of high sampling costs from multi-step denoising.

\textbf{Flow Matching and Flow Policies.} Flow matching is a generative modeling approach that learns time-varying velocity fields through regressing target vector fields, which define probability paths transporting samples from simple prior distributions to target data distributions, characterized by simple training and fast inference\citep{lipman2023flowmatching} (see Section~\ref{sec:prelim} for details). Recently, flow matching has attracted attention in reinforcement learning due to its strong expressiveness and faster inference. FQL \citep{park2025fql} uses flow matching policies to approximate behavior policies and distills flow policies for efficient policy learning and inference. FlowQ \citep{alles2025flowq} learns a conditional velocity field for flow policies by modeling energy-guided probability paths as Gaussian trajectories. However, these methods typically rely on simple Gaussian priors and multi-step sampling from flow policies, limiting representation richness and bringing additional computational overhead. To address computational bottlenecks from multi-step generative policies, MeanFlowQL \citep{wang2026meanflowql} reformulates MeanFlow \citep{geng2025meanflow} as a residual mapping that directly transforms noise into actions. SSCP \citep{koirala2026sscp} uses enhanced flow matching objective functions to predict direct completion vectors from intermediate samples, achieving one-shot action generation.\looseness=-1

Unlike the above methods, this paper introduces \ourmodel{}, achieving paradigm shifts in three dimensions. First, at the representation space level, existing flow policies operate directly in raw action space\citep{park2025fql,alles2025flowq}, requiring learning highly nonlinear mappings from Gaussian noise to complex action distributions. \ourmodel{} compresses trajectory chunks into high-level latent decision representations through an encoder, reformulating flow matching as short-range approximately linear flow mappings within decision space, thereby establishing foundations for one-step generation. Second, at the generation mechanism level, FQL\citep{park2025fql} and FlowQ\citep{alles2025flowq} learn marginal velocity fields requiring multi-step ODE integration, while \ourmodel{} learns mean velocity fields\citep{geng2025meanflow} directly predicting residual directions from starting points to endpoints, achieving one-step high-fidelity generation. Third, at the optimization objective level, existing methods indiscriminately fit dataset behavior frequencies, while \ourmodel{} introduces a dual-branch flow architecture: the Source-to-Positive Alignment Mean-Flow Branch establishes baseline mappings from Gaussian-Induced Decision Sources to dataset decisions, and the Quality-Uplift Mean-Flow Branch achieves value-guided flow mapping from suboptimal decisions to high-value decisions through hierarchical chunk-level advantage weighting, with both branches sharing the same conditional mean velocity field parameters. This design enables the model to extract flow mapping patterns leading from suboptimal trajectories to optimal policies, breaking through dataset quality ceilings. To our knowledge, this is the first offline reinforcement learning method combining mean flow matching with value-guided dual-branch flow mapping in high-level latent decision space to exploit suboptimal-to-optimal data capabilities, achieving unity of expressiveness and computational efficiency.\looseness=-1

\section{Preliminaries}
\label{sec:prelim}

\subsection{Offline Reinforcement Learning}
Reinforcement learning (RL) is typically formalized as a Markov Decision Process (MDP), defined by $\mathcal{M} = (\mathcal{S}, \mathcal{A}, P, R, \gamma)$. Here, $\mathcal{S}$ and $\mathcal{A}$ denote the state and action spaces, $P(s' \mid s, a)$ denotes the transition probability of moving from $s$ to $s'$ given action $a$, and $R(s, a)$ denotes the reward function. The discount factor $\gamma \in [0, 1)$ governs the trade-off between immediate and future rewards. In RL, the objective is to learn a policy $\pi_\theta(a \mid s)$, parameterized by $\theta$, that maximizes the expected discounted return
\begin{equation}
\mathbb{E}_\pi\left[ \sum_{h=0}^\infty \gamma^h R(s_h, a_h) \right].
\end{equation}

To support policy learning, the action-value (Q) function is defined:
\begin{equation}
Q^\pi(s, a) = \mathbb{E}_\pi\left[ \sum_{h=0}^\infty \gamma^h R(s_h, a_h) \mid s_0 = s, a_0 = a \right],
\end{equation}
which measures the expected cumulative return starting from state $s$ and action $a$ under policy $\pi$.

Offline RL is a setting of RL where the agent does not interact with the environment but instead learns from a fixed dataset of transitions $\mathcal{D} = \{(s_h, a_h, s_{h+1}, r_h)\}$. The challenge lies in learning an optimal policy solely from this static dataset that often contains suboptimal behavior, without any further exploration.

\subsection{Implicit Q-Learning}

Implicit Q-Learning (IQL)\citep{kostrikov2021offline} is an offline reinforcement learning method that learns upper-bound estimates of value functions through expectile regression, thereby avoiding out-of-distribution action evaluation. Unlike traditional Q-learning that directly fits Q-value means, IQL's core innovation lies in using asymmetric expectile loss functions to train the state-value network $V(s)$.\looseness=-1

IQL maintains two Q-networks $Q_{\omega,1}(s,a)$ and $Q_{\omega,2}(s,a)$, a state-value network $V_\nu(s)$, and corresponding target Q-networks. The value network is learned through expectile regression:
\begin{equation}
\mathcal{L}_V = \mathbb{E}\left[\left|\tau_{\mathrm{exp}}-\mathbb{I}(\delta<0)\right|\delta^2\right],
\end{equation}
where $\delta = \min(Q_{\bar\omega,1}(s,a), Q_{\bar\omega,2}(s,a)) - V_\nu(s)$, and $\tau_{\mathrm{exp}} \in (0.5, 1)$ is the expectile parameter (typically 0.7). This asymmetric loss makes $V_\nu(s)$ learn as a high expectile of the behavior value distribution rather than a simple average, thereby providing more conservative value estimates.

Based on this value baseline, an advantage function can be defined:
\begin{equation}
A(s,a) = \min(Q_{\bar\omega,1}(s,a), Q_{\bar\omega,2}(s,a)) - V_\nu(s).
\end{equation}
This advantage function can identify relatively high-value state-action pairs in the dataset, providing value guidance signals for policy learning without explicit policy improvement steps.

\subsection{Flow Matching}

Flow matching is a generative modeling approach that adopts simple and intuitive training objectives by regressing target vector fields, which define desired probability paths\citep{lipman2023flowmatching}. Flow matching learns a time-varying vector field by defining an ordinary differential equation (ODE), whose solution transports samples from a simple prior distribution to the target data distribution.

Given a data distribution $p(x)$, the goal is to learn a time-varying velocity field $v_{\theta}(x_{t}, t): \mathbb{R}^{d} \times [0,1] \to \mathbb{R}^{d}$ parameterized by $\theta$, where $x_{t} \in \mathbb{R}^{d}$ denotes a $d$-dimensional data sample and $t \in [0,1]$ denotes time. This velocity field defines a continuous flow that transports samples from a simple prior distribution (e.g., unit Gaussian) at $t=0$ to the target data distribution $p(x)$ at $t=1$.

\textbf{Flow Path Construction and Velocity Field.} Flow paths can be constructed as $x_{t}=\alpha_{t} x_{0}+\beta_{t} \epsilon$ \citep{park2025fql,lipman2023flowmatching}, where $\alpha_{t}$ and $\beta_{t}$ are predefined time-varying functions (common choice: $\alpha_t = 1-t$, $\beta_t = t$), $x_0 \sim p(x)$ is a data sample, and $\epsilon \sim \mathcal{N}(0,I)$ is noise. The corresponding \emph{conditional velocity} is given by $v_{t}(x_t \mid x_0) =\frac{d}{dt} x_{t}=\alpha_{t}' x_{0}+\beta_{t}' \epsilon$, where $'$ denotes the time derivative with respect to $t$.

Standard flow matching learns the \emph{marginal velocity field} $v(x_t,t)\triangleq\mathbb{E}_{p_t(x_0|x_t)}[v_t(x_t|x_0)]$ via the Conditional Flow Matching (CFM) loss~\citep{lipman2023flowmatching}, which requires multi-step ODE integration during generation. In contrast, the \emph{mean velocity field}\citep{geng2025mean} directly models the average velocity between two timesteps, defined as:
\begin{equation}
u(x_t, r, t) \triangleq \frac{1}{t-r} \int_r^t v(x_\tau, \tau) d\tau,
\end{equation}
where $u$ denotes the average velocity from time $r$ to time $t$, and $v$ denotes the marginal velocity. As $r \to t$, $u$ converges to $v$. The advantage of the mean velocity field is that it can directly predict residual directions from starting points to endpoints, achieving one-step high-fidelity generation without iterative ODE integration. By learning the mean velocity field $u_\theta(x_t, r, t)$, samples can be directly generated through a one-step update $x_1 = x_0 + u_\theta(x_0, 0, 1)$ over the unit time interval $\Delta t = 1$.

\textbf{Training Objective for Mean Velocity Field.} The mean velocity field $u_{\theta}$ is optimized through the mean flow identity. Defining the average velocity as $u(x_t, r, t) = \frac{1}{t-r} \int_r^t v(x_\tau, \tau) d\tau$, by taking the time derivative and applying the chain rule, the mean flow identity can be obtained\citep{geng2025mean}:
\begin{equation}
u(x_t, r, t) = v(x_t, t) - (t-r) \left( v(x_t,t) \cdot \partial_{x_t} u + \partial_t u \right),
\label{eq:meanflow_identity}
\end{equation}
where $\partial_{x_t} u$ and $\partial_t u$ denote partial derivatives of $u$ with respect to $x_t$ and $t$ respectively. This identity relates the marginal velocity $v$ to the mean velocity $u$, enabling construction of supervision targets for mean velocity through the known conditional velocity $v_t(x_t \mid x_1) = x_1 - \epsilon$. Training minimizes:
\begin{equation}
\mathcal{L}_{\mathrm{MF}}(\theta)=\mathbb{E}_{\substack{0 \le r \le t \le 1, \\ \epsilon \sim \mathcal{N}(0,I), \\ x_{1} \sim p(x)}}\left[\left\| u_{\theta}(x_{t}, r, t)-u^*(x_t, r, t \mid x_1)\right\|_{2}^{2}\right],
\label{eq:meanflow_loss}
\end{equation}
where $u^*(x_t, r, t \mid x_1)$ is computed from the conditional velocity through Equation~(\ref{eq:meanflow_identity}).

\textbf{Sampling and Generation with Mean Velocity Field.} Given the mean velocity field $u_{\theta}(x_t, r, t)$, samples can be directly generated through a one-step update:
\begin{equation}
x_1 = x_0 + u_{\theta}(x_0, 0, 1),
\label{eq:meanflow_sampling}
\end{equation}
where $x_0 = \epsilon \sim \mathcal{N}(0, I)$ is prior noise. The mean velocity field directly predicts residual directions from starting points to endpoints, achieving one-step high-fidelity generation without iterative ODE integration, significantly reducing inference costs.

\section{MeanFlow2Better Method}
\label{sec:method}

\ourmodel{} reformulates offline reinforcement learning as a quality improvement problem in high-level latent decision space. Trajectory chunks are compressed into high-level latent decision representations through a Chunk-Level Latent Decision Encoder $q_\phi$, a shared Conditional Latent Decision Mean-Flow Field $v_\theta$ learns conditional mean velocities across two supervisory branches, and a Hierarchical Action Decoder $p_\psi$ recovers continuous action sequences from refined decisions. A Hierarchical Chunk-Level Advantage Critic $\mathcal{C}_\eta$ provides value guidance during training only. Throughout this section, we fix $t_0=0$ and $t_1=1$ as the start and end of the unit flow interval. Algorithm~\ref{algo:meanflow2better} summarizes the complete pipeline.

\subsection{Chunk-Level High-Level Latent Decision Representation}
\label{subsec:latent-decision}

Given an offline dataset $\mathcal{D}=\{\tau_i\}_{i=1}^{N}$, each trajectory $\tau_i=(s_0,a_0,r_0,\ldots,s_{T-1},a_{T-1},r_{T-1})$ is segmented into consecutive chunks of length $H$:
\begin{equation}
\tau_k=(s_{k,0:H-1},\,a_{k,0:H-1}),
\end{equation}
yielding $K\approx T/H$ chunks per trajectory. The encoder $q_\phi$ maps each chunk to a distribution over high-level latent decisions:
\begin{equation}
q_\phi(\tau_k)=[\mu_k,\log\sigma_k^2]\in\mathbb{R}^{2d_z},
\end{equation}
and the full decision sequence is $Z=[z_1,\ldots,z_K]\in\mathbb{R}^{K\times 2d_z}$. The mean-flow field operates on the concatenated $[\mu,\log\sigma^2]$ representation directly. The decoder samples via reparameterization, $z_k=\mu_k+\exp(\frac{1}{2}\log\sigma_k^2)\odot\epsilon_k$, and generates actions as $\hat{a}_{k,0:H-1}=p_\psi(s_{k,0:H-1},z_k)$. This separation of high-level behavioral intention from low-level action execution allows the flow model to perform behavioral improvement in a compact, stable decision space rather than over raw action sequences.

\subsection{High-Level Decision Source Construction}
\label{subsec:decision-source}

\subsubsection{Gaussian-Induced Decision Source}

A Gaussian pseudo-trajectory $\tau^-_G\sim\mathcal{N}(0,I)$ is drawn with the same temporal shape as real trajectories. \textbf{Inspired by the neural null-space mechanism}~\citep{STAVISKY2017195}, we \textbf{explicitly set its reward, RTG, chunk-level RTG, and trajectory return to zero}, placing Gaussian-Induced Decision Sources at the zero point of the value dimension. After encoding:
\begin{equation}
Z_{\mathrm{neg}}=q_\phi(\tau^-_G).
\end{equation}
The Source-to-Positive Alignment Mean-Flow Branch transports $Z_{\mathrm{neg}}$ toward positive behavioral decision regions, establishing task-relevant structure from a value-neutral baseline unconstrained by any specific return interval in the dataset.\looseness=-1

\subsubsection{Suboptimal-to-High-Value Trajectory Pairing}

Trajectories are partitioned into a high-value pool $\mathcal{I}_{\mathrm{top}}$ and a suboptimal pool $\mathcal{I}_{\mathrm{low}}$ by sorting on complete return $R_i=\sum_t r_{i,t}$. The high-value timestep coverage ratio $\rho_{\mathrm{traj}}\in(0,1]$ controls the timestep budget of $\mathcal{I}_{\mathrm{top}}$. Suboptimal and high-value latent decisions are then obtained by encoding trajectories sampled from each pool:
\begin{equation}
Z_{\mathrm{low}}=q_\phi(\tau_{\mathrm{low}}),
\qquad
Z_{\mathrm{top}}=q_\phi(\tau_{\mathrm{top}}).
\end{equation}
The Quality-Uplift Mean-Flow Branch learns to transport $Z_{\mathrm{low}}$ toward $Z_{\mathrm{top}}$, distilling high-value decision structure from existing suboptimal experiences.

\subsection{Dual-Branch Conditional Latent Decision Mean Flow Mapping}
\label{subsec:dual-branch-flow}

Both branches share the Conditional Latent Decision Mean-Flow Field $v_\theta(Z, t_0, t_1, c)$ but differ in source, target, and condition. Return-sorted pairing ensures the model learns mean velocities aligned with return improvement rather than arbitrary intra-distribution variation.

Each branch receives chunk-level and global conditions. The local condition for the $k$-th chunk is the target trajectory's return-to-go:
\begin{equation}
c_{\mathrm{chunk}}^{(b,k)}=\sum_{t=kH}^{T-1}r_{\mathrm{top},t}^{(b)},
\end{equation}
and the global condition is the return gap:
\begin{equation}
c_{\mathrm{global}}^{(b)}=R_{\mathrm{top}}^{(b)}-R_{\mathrm{low}}^{(b)}.
\end{equation}
For the Source-to-Positive Alignment branch, the pair is $(\tau_{\mathrm{neg}},\tau_{\mathrm{pos}})$ with $R_{\mathrm{neg}}=0$, so $c_{\mathrm{global}}=R_{\mathrm{pos}}^{(b)}$. Both conditions are independently projected and injected into chunk tokens and a global condition token.

\subsubsection{Source-to-Positive Alignment Mean-Flow Branch}

The branch predicts the mean velocity transporting $Z_{\mathrm{neg}}$ to $Z_{\mathrm{pos}}$ over the unit interval $[t_0, t_1]$:
\begin{equation}
\widehat{U}_{\mathrm{flow}}^{(b)}
=v_\theta\!\left(Z_{\mathrm{neg}}^{(b)},\,t_0,\,t_1,\,c_{\mathrm{flow}}^{(b)}\right),
\qquad
U_{\mathrm{flow}}^{*(b)}=\frac{Z_{\mathrm{pos}}^{(b)}-Z_{\mathrm{neg}}^{(b)}}{t_1-t_0}.
\end{equation}

\subsubsection{Quality-Uplift Mean-Flow Branch}

The branch predicts the mean velocity transporting $Z_{\mathrm{low}}$ to $Z_{\mathrm{top}}$ over the unit interval $[t_0, t_1]$:
\begin{equation}
\widehat{U}_{\mathrm{main}}^{(b)}
=v_\theta\!\left(Z_{\mathrm{low}}^{(b)},\,t_0,\,t_1,\,c_{\mathrm{main}}^{(b)}\right),
\qquad
U_{\mathrm{main}}^{*(b)}=\frac{Z_{\mathrm{top}}^{(b)}-Z_{\mathrm{low}}^{(b)}}{t_1-t_0}.
\end{equation}

Refined high-level decisions are obtained by one-step integration over $[t_0, t_1]$:
\begin{equation}
Z_{\mathrm{plus}}^{(b)}=Z_{\mathrm{low}}^{(b)}+(t_1-t_0)\,\widehat{U}_{\mathrm{main}}^{(b)},
\end{equation}
which are decoded to reconstruct action sequences for the target high-value trajectory. This branch directly instantiates the core hypothesis: learnable improvement directions exist between suboptimal and high-value behaviors in high-level decision space.

\subsubsection{High-Level Decision Transformer Architecture}

The mean-flow field $v_\theta$ is a Transformer with bidirectional multi-head self-attention over the chunk sequence, taking $Z\in\mathbb{R}^{K\times 2d_z}$ as input tokens. Unlike the causal encoder and decoder, the flow field uses full attention so every chunk token can attend to the complete high-level decision sequence, enabling globally coherent mean velocity prediction. The attention sequence length $K\approx T/H$ compresses the original $T$-step horizon into $K$ decision units.

\subsection{Hierarchical Chunk-Level Latent Decision Value Estimation and Advantage Guidance}
\label{subsec:chunk-critic}

To avoid indiscriminate fitting to dataset behavior frequency, a hierarchical chunk-level advantage critic provides value guidance. Following IQL~\citep{kostrikov2021offline} (see Section~\ref{sec:prelim}), the critic operates on chunk-level transitions $(s_{b,k}, z_{b,k}, r_{b,k}^{\mathrm{chunk}}, s_{b,k+1})$ rather than primitive actions or full trajectories, and is used only during training.

The Bellman target is:
\begin{equation}
y_{b,k}=r_{b,k}^{\mathrm{chunk}}+\gamma_{\mathrm{boot}}\,V_\nu(s_{b,k+1}),
\end{equation}
with $\gamma_{\mathrm{boot}}=0.99$ and target networks maintained via Polyak averaging. The expectile value loss learns $V_\nu(s)$ as a high-expectile baseline of the behavioral decision value distribution. Chunk-level advantage is:
\begin{equation}
A_{b,k}=\min\!\left(Q_{\bar\omega,1}(s_{b,k},z_{b,k}),\,Q_{\bar\omega,2}(s_{b,k},z_{b,k})\right)-V_\nu(s_{b,k}).
\end{equation}

Advantages are converted to exponential weights:
\begin{equation}
w_{b,k}\propto\exp(\beta\,A_{b,k}).
\end{equation}
The Quality-Uplift branch uses a weight interpolated between $w_{b,k}$ and uniform to prevent extreme chunks from dominating flow supervision while preserving quality preference. The Source-to-Positive branch uses uniform weighting. Low-value chunks remain as source endpoints for the Quality-Uplift branch. The critic adjusts their training contribution rather than discarding them.

\subsection{Joint Training Objective}
\label{subsec:training-objective}

The total loss is:
\begin{equation}
\mathcal{L}
=\lambda_{\mathrm{act}}\mathcal{L}_{\mathrm{act}}
+\lambda_{\mathrm{main}}\mathcal{L}_{\mathrm{main}}
+\lambda_{\mathrm{flow}}\mathcal{L}_{\mathrm{flow}}.
\end{equation}

The action reconstruction loss reweights samples by hierarchical chunk-level advantage:
\begin{equation}
\mathcal{L}_{\mathrm{act}}
=\mathbb{E}_{b,k,t}\!\left[w_{b,k}\left\|\hat{a}_{bkt}-a_{bkt}^{\mathrm{top}}\right\|_2^2\right].
\end{equation}

To prevent flow supervision targets from drifting jointly with encoder parameters, stop-gradient is applied to both endpoints when constructing mean velocity targets. Let $\operatorname{sg}(\cdot)$ denote the stop-gradient operator:
\begin{equation}
U_{\mathrm{main}}^*
=\frac{\operatorname{sg}(Z_{\mathrm{top}})-\operatorname{sg}(Z_{\mathrm{low}})}{t_1-t_0},
\qquad
U_{\mathrm{flow}}^*
=\frac{\operatorname{sg}(Z_{\mathrm{pos}})-\operatorname{sg}(Z_{\mathrm{neg}})}{t_1-t_0}.
\end{equation}
This ensures $\mathcal{L}_{\mathrm{main}}$ and $\mathcal{L}_{\mathrm{flow}}$ update $v_\theta$ only through the predicted velocity branch and do not backpropagate into the encoder via the target construction path. The encoder receives end-to-end gradients through $\mathcal{L}_{\mathrm{act}}$ and the decoding path from $Z_{\mathrm{plus}}$.

The flow losses are normalized by the target velocity scale:
\begin{equation}
\mathcal{L}_{\mathrm{main}}
=\frac{\mathbb{E}_{b,k}\!\left[w_{b,k}^{\mathrm{main}}\left\|\widehat{U}_{\mathrm{main}}-U_{\mathrm{main}}^*\right\|_2^2\right]}{\mathbb{E}\!\left[\left\|U_{\mathrm{main}}^*\right\|_2^2\right]},
\qquad
\mathcal{L}_{\mathrm{flow}}
=\frac{\mathbb{E}_{b,k}\!\left[\left\|\widehat{U}_{\mathrm{flow}}-U_{\mathrm{flow}}^*\right\|_2^2\right]}{\mathbb{E}\!\left[\left\|U_{\mathrm{flow}}^*\right\|_2^2\right]}.
\end{equation}

\begin{algorithm}[t]
\caption{\ourmodel{} Training and Inference}
\label{algo:meanflow2better}
\begin{algorithmic}[1]
\STATE \textbf{Input:} Offline dataset $\mathcal{D}$, coverage ratio $\rho_{\mathrm{traj}}$; fix $t_0=0,\,t_1=1$
\STATE \textbf{Initialize:} $q_\phi$, $v_\theta$, $p_\psi$, $Q_{\omega,1}$, $Q_{\omega,2}$, $V_\nu$
\STATE Sort by return; partition $\mathcal{I}_{\mathrm{top}}$ and $\mathcal{I}_{\mathrm{low}}$
\WHILE{not converged}
    \STATE Sample $\tau_{\mathrm{low}}\!\sim\!\mathcal{I}_{\mathrm{low}}$,\; $\tau_{\mathrm{top}}\!\sim\!\mathcal{I}_{\mathrm{top}}$,\; $\tau_{\mathrm{neg}}\!\sim\!\mathcal{N}(0,I)$,\; $\tau_{\mathrm{pos}}\!\sim\!\mathcal{D}$
    \STATE Encode: $Z_{\mathrm{low}},Z_{\mathrm{top}},Z_{\mathrm{neg}},Z_{\mathrm{pos}}\leftarrow q_\phi(\cdot)$
    \STATE Compute $A_{b,k}$, $w_{b,k}$ via hierarchical chunk-level IQL critic
    \STATE $\widehat{U}_{\mathrm{flow}}\leftarrow v_\theta(Z_{\mathrm{neg}},t_0,t_1,c_{\mathrm{flow}})$;\quad $U^*_{\mathrm{flow}}\leftarrow\bigl(\operatorname{sg}(Z_{\mathrm{pos}})-\operatorname{sg}(Z_{\mathrm{neg}})\bigr)/(t_1-t_0)$
    \STATE $\widehat{U}_{\mathrm{main}}\leftarrow v_\theta(Z_{\mathrm{low}},t_0,t_1,c_{\mathrm{main}})$;\quad $U^*_{\mathrm{main}}\leftarrow\bigl(\operatorname{sg}(Z_{\mathrm{top}})-\operatorname{sg}(Z_{\mathrm{low}})\bigr)/(t_1-t_0)$
    \STATE $Z_{\mathrm{plus}}\leftarrow Z_{\mathrm{low}}+(t_1-t_0)\,\widehat{U}_{\mathrm{main}}$;\quad $\hat{A}_{\mathrm{top}}\leftarrow p_\psi(S_{\mathrm{top}},Z_{\mathrm{plus}})$
    \STATE Update $\theta,\phi,\psi$: minimize $\lambda_{\mathrm{act}}\mathcal{L}_{\mathrm{act}}+\lambda_{\mathrm{main}}\mathcal{L}_{\mathrm{main}}+\lambda_{\mathrm{flow}}\mathcal{L}_{\mathrm{flow}}$
    \STATE Update $Q_{\omega,1},Q_{\omega,2},V_\nu$; Polyak-update target networks
\ENDWHILE
\STATE \textbf{// Inference: Online Decision Execution}
\STATE \textbf{Input:} Trained $q_\phi$, $v_\theta$, $p_\psi$; target return condition $c$
\FOR{each episode}
    \STATE Reset environment; initialize observation buffer $\mathcal{B}\leftarrow\{s_0\}$
    \WHILE{episode not done}
        \STATE Construct source chunk $\tau_{\mathrm{source}}\leftarrow(s_{t-H+1:t},\,a_{t-H+1:t-1})$ from buffer $\mathcal{B}$
        \STATE $Z_{t_0}\leftarrow q_\phi(\tau_{\mathrm{source}})$ \hfill\COMMENT{encode to high-level latent decision}
        \STATE $Z_{t_1}\leftarrow Z_{t_0}+(t_1-t_0)\,v_\theta(Z_{t_0},t_0,t_1,c)$ \hfill\COMMENT{one-step mean-flow refinement}
        \STATE $\hat{a}_{0:H-1}\leftarrow p_\psi(s_{t-H+1:t},\,Z_{t_1})$ \hfill\COMMENT{decode $H$ actions}
        \FOR{$h=0$ \TO $H-1$}
            \STATE Execute $\hat{a}_h$ in environment; observe $s_{t+h+1}$, $r_{t+h}$
        \ENDFOR
        \STATE Update buffer $\mathcal{B}$ with new states $s_{t+1:t+H}$;\quad $t\leftarrow t+H$
    \ENDWHILE
\ENDFOR
\end{algorithmic}
\end{algorithm}

\subsection{One-Step High-Level Decision Refinement and Inference}
\label{subsec:inference}

Given a source trajectory at test time, inference reduces to:
\begin{align}
Z_{t_0} &= q_\phi(\tau_{\mathrm{source}}), \\
Z_{t_1} &= Z_{t_0} + (t_1-t_0)\,v_\theta(Z_{t_0},\,t_0,\,t_1,\,c), \\
\hat{a}_{0:T-1} &= p_\psi(Z_{t_1},\,s_{0:T-1}).
\end{align}
The mean-flow field performs a single forward pass to predict the full mean velocity over $[t_0,t_1]$, so no iterative ODE integration is required. The core of the method is not generating trajectories from scratch, but learning directional transport from existing suboptimal behaviors toward high-value behavioral regions in structured high-level decision space.\looseness=-1

\section{Experiments}
\label{sec:exp}

\subsection{Environmental Setup}

We evaluate \ourmodel{} on 9 Gym-MuJoCo locomotion tasks from the D4RL benchmark~\citep{fu2020d4rl}, covering HalfCheetah, Hopper, and Walker2d under medium, medium-replay, and medium-expert dataset conditions.

\subsection{Evaluation Metrics}

We report D4RL normalized scores: $\text{Normalized Score} = 100 \times (\text{score} - \text{random score})/(\text{expert score} - \text{random score})$, where 0 corresponds to a uniform random policy and 100 to a domain-specific expert. For evaluation, we report the average normalized return over 5 training seeds, with each model evaluated for 50 episodes per task.
\subsection{Baseline Models}

We compare \ourmodel{} with representative baselines across three categories: \textbf{regression-based policies} (BC, TD3-BC~\citep{fujimoto2021minimalist}, IQL~\citep{kostrikov2021offline}, CQL~\citep{kumar2020conservative}, MOReL~\citep{kidambi2020morel}, ReBRAC~\citep{tarasov2023rebrac}, SRPO~\citep{chen2024srpo}, CAC~\citep{zhang2021cautious}, FAMAC, FERAC), \textbf{diffusion policies} (Diffuser~\citep{janner2022planning}, DD~\citep{ajay2022conditional}, EDP~\citep{kang2023edp}, IDQL~\citep{hansen2023idql}, DQL~\citep{wang2022diffusion}, TCD~\citep{hu2026tcd}), and \textbf{flow policies} (FQL~\citep{park2025fql}, IFQL, FlowQ~\citep{alles2025flowq}, LaMPO~\citep{11641025}, SSCP~\citep{koirala2026sscp}, Value Flows~\citep{dong2025vf}, FAN~\citep{lee2026efficientexpressiveofflinerl}).

\section{Experimental Results}
\label{sec:results}

\subsection{Main Performance Comparison}

\subsubsection{Gym-MuJoCo Locomotion Task Comprehensive Comparison}

Table \ref{tab:gym_comprehensive} presents the comprehensive comparison of \ourmodel{} with existing offline RL methods on 9 Gym-MuJoCo tasks. We categorize baseline models into three classes: regression-based policies, diffusion policies, and one-step flow policies. \ourmodel{} achieves an average score of 90.91 and total score of 818.22 on Gym9 tasks, performing particularly well on medium-replay datasets, which contain suboptimal and noisy trajectories posing challenges for standard policy learning. It achieves competitive performance among one-step flow policies.

\begin{table*}[t]
\scriptsize
\centering
\caption{D4RL Gym-MuJoCo Benchmark Comprehensive Comparison.}
\label{tab:gym_comprehensive}
\setlength{\tabcolsep}{4.5pt}
\resizebox{\textwidth}{!}{
\begin{tabular}{l|cccccc|cccccc|cccc}
\toprule
& \multicolumn{6}{c|}{\textbf{Regression-based Policies}} & \multicolumn{6}{c|}{\textbf{Diffusion Policies}} & \multicolumn{4}{c}{\textbf{Flow Policies}} \\
\midrule
Dataset & BC & TD3-BC & IQL & CQL & MOReL & CAC & Diffuser & DD & EDP & IDQL & DQL & TCD & FQL & FlowQ & SSCP & \textbf{\ourmodel{}} \\
\midrule
HalfCheetah ME & 55.2 & 90.7 & 86.7 & 91.6 & 53.3 & 84.3 & 90.3±0.1 & 88.9±1.9 & 95.8±0.1 & 91.3±0.6 & 96.8±0.3 & 92.7±3.4 & \textbf{99.8±0.1} & 89.6±2.3 & 98.1±0.9 & 95.47±0.31 \\
Hopper ME & 52.5 & 98.0 & 91.5 & 105.4 & 108.7 & 100.4 & 107.2±0.9 & 110.4±0.6 & 110.8±0.4 & 110.1±0.7 & 111.1±1.3 & 112.6±1.0 & 86.2±1.3 & 103.5±7.0 & 110.9±0.9 & \textbf{119.74±1.24} \\
Walker2d ME & 107.5 & 110.1 & 109.6 & 108.8 & 95.6 & 110.4 & 107.4±0.1 & 108.4±0.1 & 110.4±0.0 & 110.6±0.0 & 110.1±0.3 & 111.3±0.7 & 100.5±0.1 & 109.1±0.2 & 111.1±0.1 & \textbf{117.78±0.95} \\
\midrule
HalfCheetah M & 42.6 & 48.3 & 47.4 & 44.0 & 42.1 & 69.1 & 43.8±0.1 & 45.3±0.3 & 50.8±0.0 & 51.5±0.1 & 51.1±0.5 & 47.2±0.7 & \textbf{60.1±0.1} & 56.8±0.8 & 52.3±0.5 & 46.69±0.38 \\
Hopper M & 52.9 & 59.3 & 66.3 & 58.5 & 95.4 & 80.7 & 89.5±0.7 & 98.2±0.1 & 72.6±0.2 & 70.1±2.0 & 90.5±4.6 & 99.4±0.6 & 74.5±0.2 & 88.9±6.4 & \textbf{102.4±0.2} & 100.93±0.89 \\
Walker2d M & 75.3 & 83.7 & 78.3 & 72.5 & 77.8 & 83.1 & 79.4±1.0 & 79.6±0.9 & 86.5±0.2 & 88.1±0.4 & 87.0±0.9 & 82.1±1.8 & 72.7±0.8 & 84.9±0.4 & 84.2±0.9 & \textbf{93.20±1.44} \\
\midrule
HalfCheetah MR & 36.6 & 44.6 & 44.2 & 45.5 & 40.2 & 58.7 & 36.0±0.7 & 42.9±0.1 & 44.9±0.4 & 46.5±0.3 & 47.8±0.3 & 40.6±1.4 & \textbf{51.1±0.1} & 50.3±0.9 & 44.4±1.0 & 43.81±0.30 \\
Hopper MR & 18.1 & 60.9 & 94.7 & 95.0 & 93.6 & 99.7 & 91.8±0.5 & 99.2±0.2 & 83.0±1.7 & 99.4±0.1 & 101.3±0.6 & 97.2±2.4 & 85.4±0.5 & 99.2±1.4 & 101.4±0.4 & \textbf{104.36±2.12} \\
Walker2d MR & 26.0 & 81.8 & 73.9 & 77.2 & 49.8 & 79.5 & 58.3±1.8 & 75.6±0.6 & 87.0±2.6 & 89.1±2.4 & 95.5±1.5 & 88.0±1.9 & 82.1±1.2 & 88.5±2.9 & 85.9±13.4 & \textbf{96.24±1.02} \\
\midrule
\textbf{Total (Gym9)} & 466.7 & 677.4 & 692.6 & 698.5 & 656.5 & 765.9 & 703.7 & 748.5 & 741.8 & 756.7 & 791.2 & 771.1 & 712.3 & 770.8 & 788.3 & \textbf{818.22} \\
\textbf{Average (Gym9)} & 51.9 & 75.3 & 77.0 & 77.6 & 72.9 & 85.1 & 78.2 & 83.2 & 82.4 & 84.1 & 87.9 & 85.7 & 79.2 & 85.6 & 87.9 & \textbf{90.91} \\
\bottomrule
\end{tabular}
}
\end{table*}

From Table~\ref{tab:gym_comprehensive}, \ourmodel{} achieves an average score of 90.91 on Gym9, surpassing all flow policies (FQL 79.2, FlowQ 85.6, SSCP 87.9) and most diffusion policies (DQL 87.9, TCD 85.7), demonstrating the efficiency of one-step conditional mean velocity field flow mapping on continuous control tasks.

\section{Ablation Studies}
\label{sec:ablation}

\subsection{Module Ablation Experiments}

Module ablations are conducted on 9 Gym-MuJoCo tasks. Each variant removes or replaces exactly one mechanism while keeping all other settings identical to the complete model.

\paragraph{Direct High-Level Latent Decision Mapping (MF2B-DLM)} Replaces conditional mean velocity transport with a direct mapping $\hat{Z}_{\mathrm{top}}=f_{\mathrm{DLM}}(Z_{\mathrm{low}},c_{\mathrm{chunk}},c_{\mathrm{global}})$, testing whether the mean velocity flow formulation is necessary.

\paragraph{No Value Guidance (MF2B-NVG)} Removes hierarchical chunk-level IQL advantage estimation and reweighting, testing whether value guidance is required to steer high-level latent decisions toward high-value regions.

\paragraph{Return-Ordered Uniform Pairing (MF2B-UP)} Draws trajectory pairs uniformly from the full dataset while preserving return direction ($R_{\mathrm{top}}\geq R_{\mathrm{low}}$), removing the explicit high-value and suboptimal trajectory pools. Tests whether directional pool-based pairing provides supervision beyond simple return-ordered pairs.

\paragraph{No Source-to-Positive Alignment Mean-Flow Branch (MF2B-NAF)} Sets the Source-to-Positive Alignment branch loss weight to zero, removing auxiliary supervision from Gaussian-Induced Decision Sources ($Z_{\mathrm{neg}}$) to positive behavioral decisions ($Z_{\mathrm{pos}}$) while retaining the Quality-Uplift Mean-Flow Branch. Tests whether the structure-forming branch is necessary for establishing task-relevant behavioral structure in the high-level latent decision space.

\subsubsection{Module Ablation Experiment Results}

Table~\ref{tab:ablation_results} presents results on Gym-MuJoCo9. The complete \ourmodel{} (average 90.91) outperforms all variants. Key findings: (1) MF2B-DLM ($-$4.47) confirms the necessity of conditional mean velocity transport over direct target mapping. (2) MF2B-NVG ($-$4.07) shows chunk-level value guidance is critical for directing latent decisions toward high-value regions. (3) MF2B-UP ($-$5.71) demonstrates that explicit directional pool-based pairing provides stronger supervision than uniform return-ordered pairs. (4) \textbf{MF2B-NAF ($-$5.28)} validates that the Source-to-Positive Alignment Mean-Flow Branch—analogous to the neural null-space mechanism~\citep{STAVISKY2017195}—establishes task-relevant topological structures and value gradient baselines that support subsequent quality-uplift transport.

\begin{table*}[htbp]
\centering
\caption{Module Ablation Experiment Results (Gym-MuJoCo 9 Tasks)}
\label{tab:ablation_results}
\resizebox{\textwidth}{!}{
\begin{tabular}{lrrrr|r}
\hline
Dataset & MF2B-DLM & MF2B-NVG & MF2B-UP & MF2B-NAF & \textbf{\ourmodel{}} \\
\hline
HalfCheetah Medium & 45.12 ± 0.25 & 46.09 ± 0.65 & 45.61 ± 0.38 & 44.99 ± 0.13 & \textbf{46.69 ± 0.38} \\
HalfCheetah Medium-Replay & 43.11 ± 0.42 & 42.33 ± 0.19 & 42.97 ± 0.26 & 42.37 ± 0.12 & \textbf{43.81 ± 0.30} \\
HalfCheetah Medium-Expert & 94.01 ± 0.22 & 94.25 ± 0.17 & 94.34 ± 0.31 & 94.74 ± 0.49 & \textbf{95.47 ± 0.31} \\
Hopper Medium & 81.57 ± 2.23 & 90.91 ± 3.83 & 92.74 ± 7.14 & 77.46 ± 1.17 & \textbf{100.93 ± 0.89} \\
Hopper Medium-Replay & 99.97 ± 3.37 & 93.10 ± 1.80 & 83.00 ± 13.69 & 97.28 ± 2.30 & \textbf{104.36 ± 2.12} \\
Hopper Medium-Expert & 117.75 ± 1.39 & 116.92 ± 0.72 & 116.71 ± 0.39 & 115.13 ± 0.51 & \textbf{119.74 ± 1.24} \\
Walker2d Medium & 88.37 ± 0.87 & 88.16 ± 0.43 & 86.15 ± 0.68 & 87.10 ± 0.35 & \textbf{93.20 ± 1.44} \\
Walker2d Medium-Replay & 89.94 ± 0.32 & 92.10 ± 0.66 & 90.76 ± 4.39 & 91.92 ± 1.42 & \textbf{96.24 ± 1.02} \\
Walker2d Medium-Expert & 114.74 ± 1.27 & 114.26 ± 1.37 & 111.12 ± 0.60 & 111.30 ± 0.20 & \textbf{117.78 ± 0.95} \\
\hline
\textbf{Total (Gym9)} & 774.58 & 778.12 & 763.40 & 767.29 & \textbf{818.22} \\
\textbf{Average (Gym9)} & 86.06 & 86.46 & 84.82 & 85.25 & \textbf{90.91} \\
\hline
\end{tabular}
}
\end{table*}

\subsubsection{High-Level Latent Decision Space Distribution Analysis}
\label{subsec:latent-diagnostic}

To understand the geometric structure of the learned high-level latent decision space, we analyze four types of representations: low-return behavioral latent decisions ($Z_{\mathrm{low}}$), high-return behavioral latent decisions ($Z_{\mathrm{top}}$), Gaussian source negative endpoints ($Z_{\mathrm{neg}}$), and source-to-positive positive endpoints ($Z_{\mathrm{pos}}$). Note that all analyses operate on the model's high-level latent representations, not the original action space.

\begin{figure*}[htbp]
    \centering
    \begin{minipage}[c]{0.30\textwidth}
        \centering
        \includegraphics[width=\linewidth]{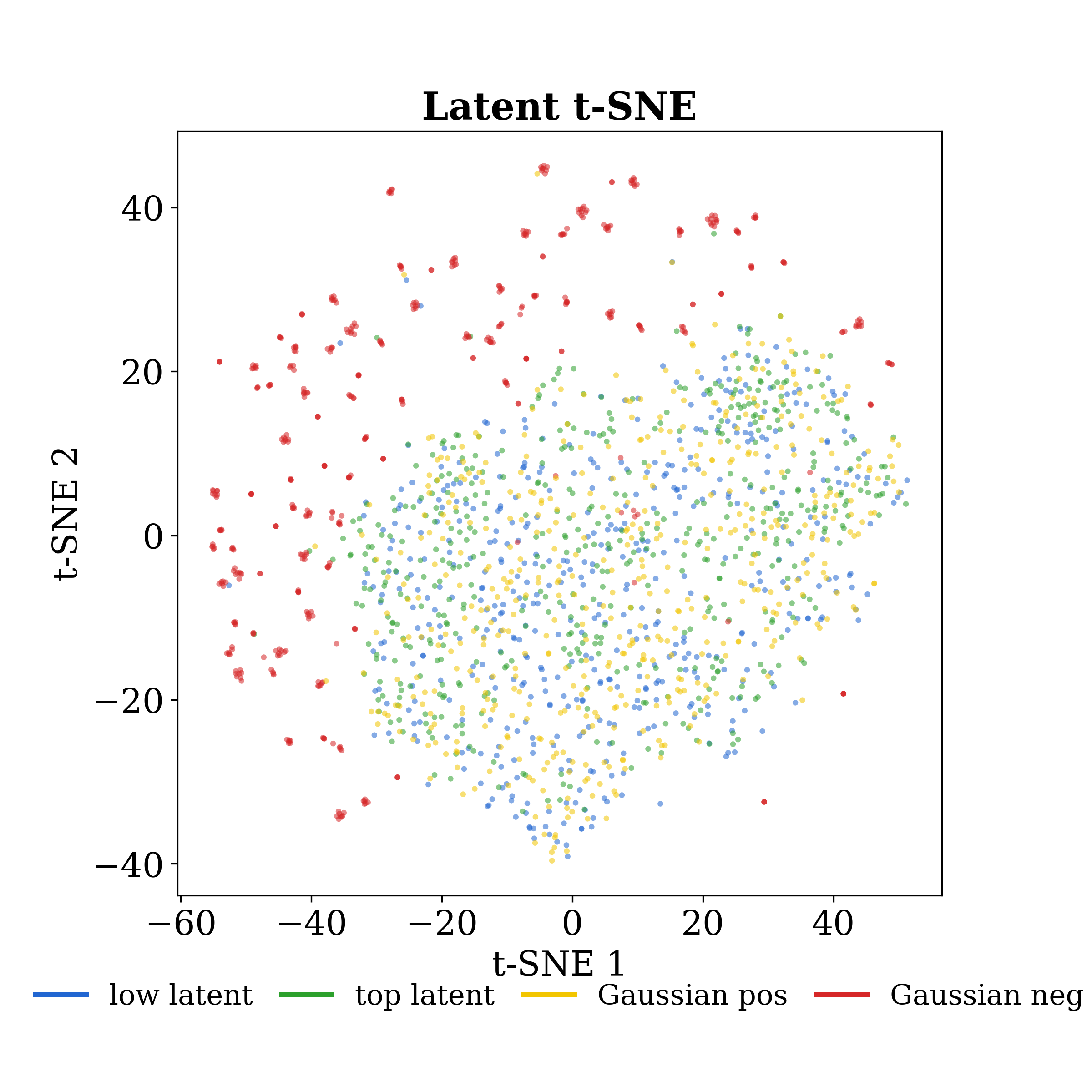}
        \subcaption{t-SNE visualization of four types of high-level latent decision representations. $Z_{\mathrm{pos}}$ closely overlaps with $Z_{\mathrm{top}}$. $Z_{\mathrm{neg}}$ forms an independent cluster. $Z_{\mathrm{low}}$ retains partial neighborhood continuity with $Z_{\mathrm{top}}$.}
        \label{fig:latent_tsne}
    \end{minipage}
    \hfill
    \begin{minipage}[t]{0.30\textwidth}
        \centering
        \includegraphics[width=\linewidth]{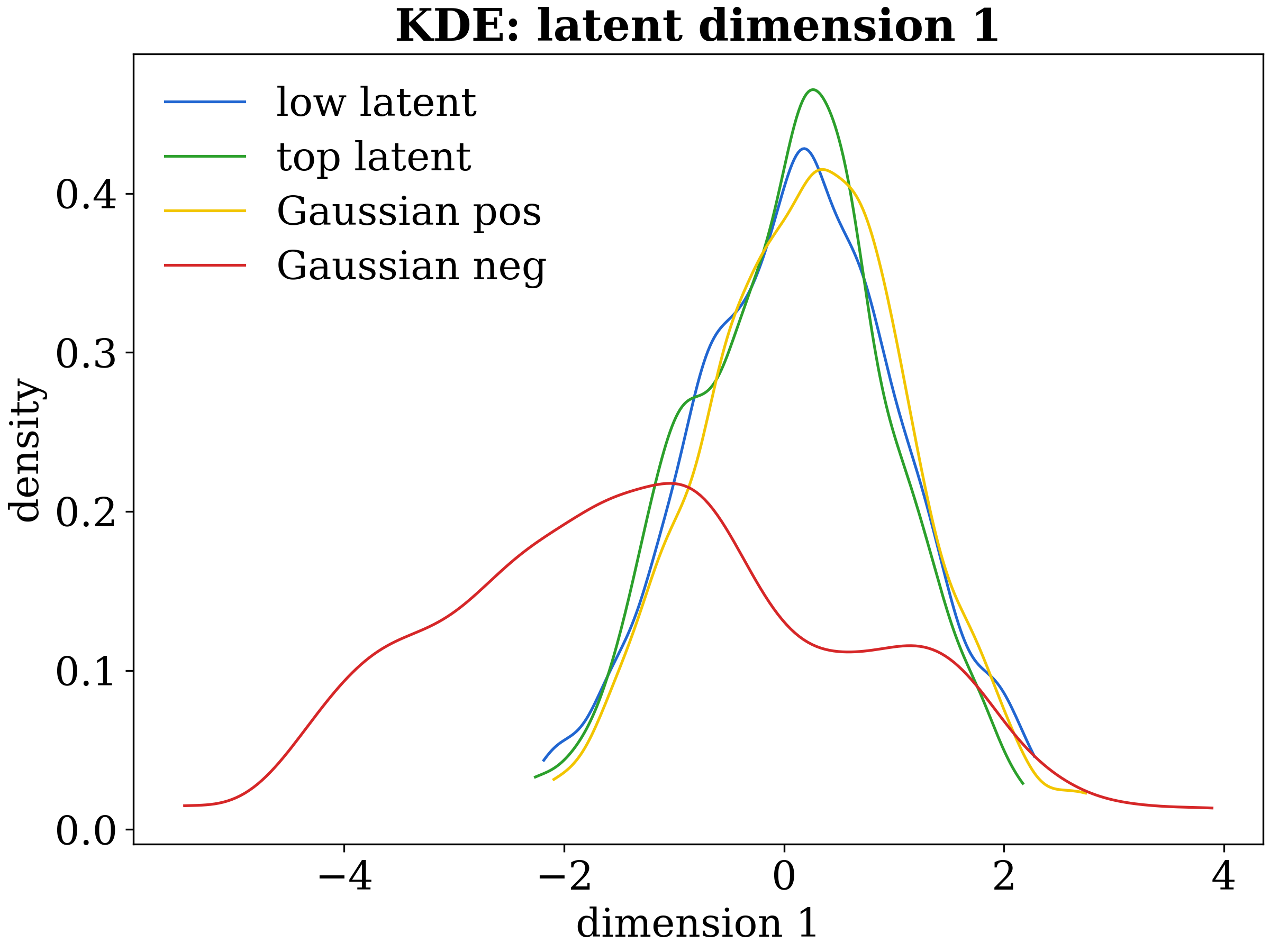}
        \vspace{0.3em}
        \includegraphics[width=\linewidth]{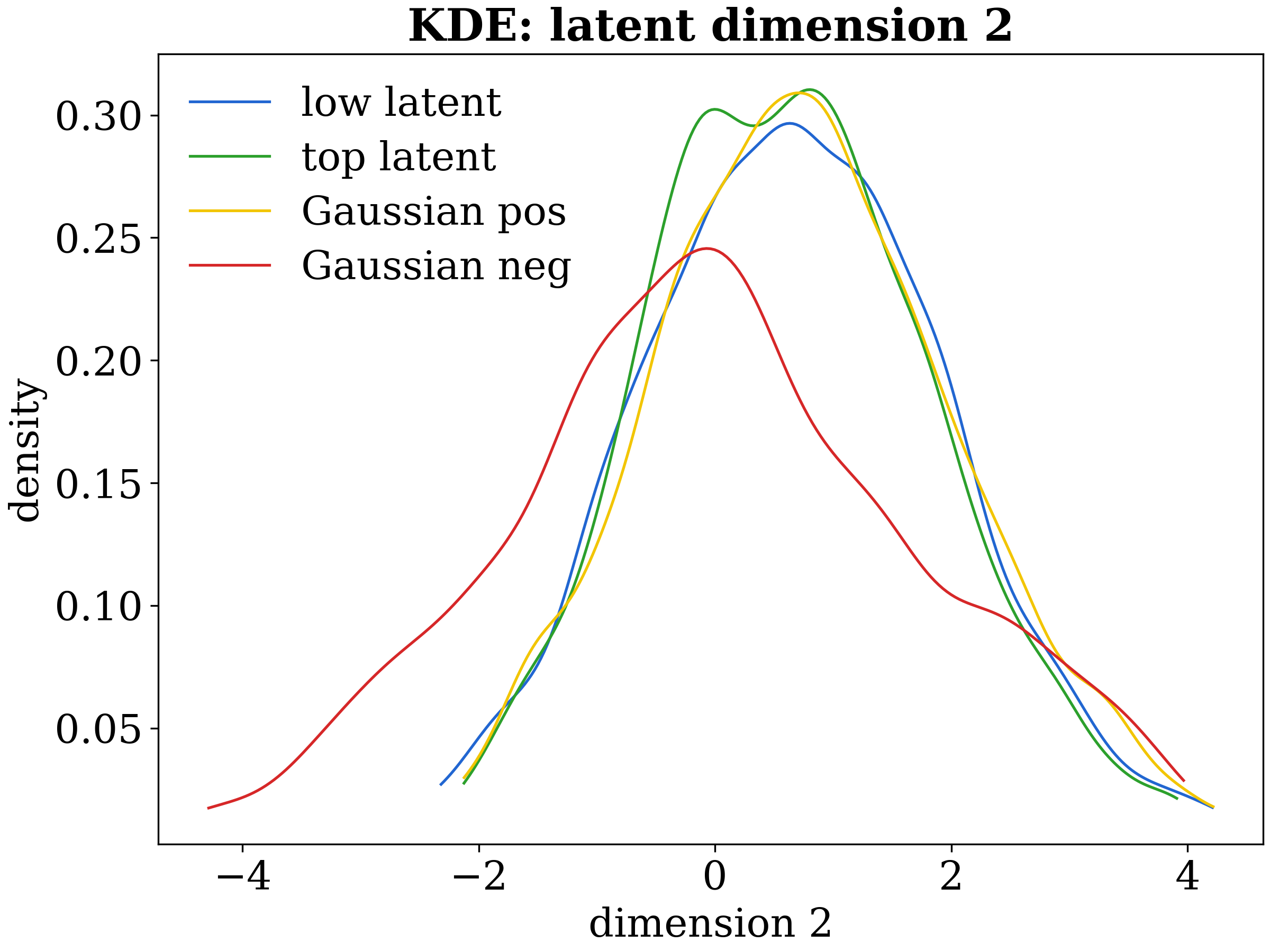}
    \end{minipage}
    \hfill
    \begin{minipage}[t]{0.30\textwidth}
        \centering
        \includegraphics[width=\linewidth]{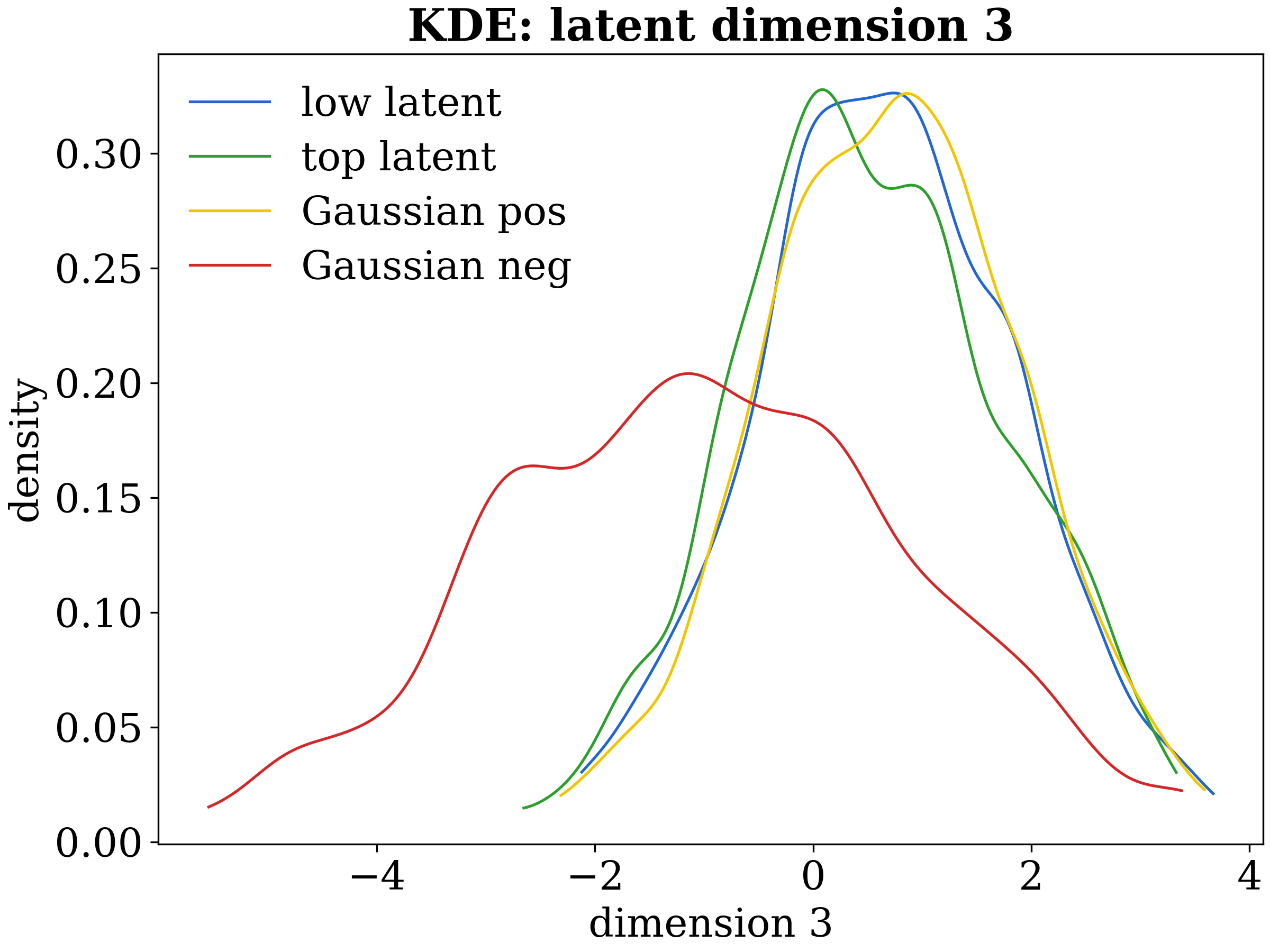}
        \vspace{0.5em}
        \subcaption{\small KDE plots of the first three latent dimensions. $Z_{\mathrm{pos}}$ closely follows $Z_{\mathrm{top}}$. $Z_{\mathrm{neg}}$ significantly deviates. $Z_{\mathrm{low}}$ retains density overlap with $Z_{\mathrm{top}}$ in major regions.}
        \label{fig:latent_kde_top_low_pos_neg}
    \end{minipage}
    \caption{High-level latent decision space analysis. Left: t-SNE visualization of $Z_{\mathrm{top}}$ (green), $Z_{\mathrm{low}}$ (blue), $Z_{\mathrm{pos}}$ (yellow), and $Z_{\mathrm{neg}}$ (red). Right: KDE plots of the first three latent dimensions comparing the four distributions.}
    \label{fig:latent_analysis}
\end{figure*}

Figure~\ref{fig:latent_tsne} shows t-SNE visualizations of the four distributions. $Z_{\mathrm{pos}}$ closely overlaps with $Z_{\mathrm{top}}$, confirming that the Source-to-Positive Alignment branch targets align with high-value behavioral regions. $Z_{\mathrm{neg}}$ forms a clearly separated cluster, demonstrating that Gaussian-Induced Decision Sources provide non-trivial structure-forming training signals rather than copying the behavioral distribution. $Z_{\mathrm{low}}$ and $Z_{\mathrm{top}}$ maintain partial neighborhood continuity, supporting the shared high-level latent decision space hypothesis and confirming that the Quality-Uplift branch performs directional improvement within a common representation space rather than cross-domain generation. KDE analyses in Figure~\ref{fig:latent_kde_top_low_pos_neg} quantitatively corroborate these findings across the first three latent dimensions.

The KDE plots in Figure~\ref{fig:latent_kde_top_low_pos_neg} reveal three quantitative patterns across the first three latent dimensions. First, $Z_{\mathrm{pos}}$ and $Z_{\mathrm{top}}$ exhibit nearly identical density curves in all three dimensions—their peaks, scales, and tail shapes closely coincide—confirming that the Source-to-Positive Alignment branch successfully transports Gaussian sources into the high-value behavioral region. Second, $Z_{\mathrm{neg}}$ deviates significantly from $Z_{\mathrm{top}}$ in location, scale, and shape, with particularly pronounced differences in dimensions 1 and 3 where $Z_{\mathrm{neg}}$ extends further in the negative direction with heavier tails, validating that Gaussian-Induced Decision Sources occupy a semantically neutral region distinct from task-relevant behaviors. Third, $Z_{\mathrm{low}}$ retains substantial density overlap with $Z_{\mathrm{top}}$ in the major probability mass regions of all three dimensions, confirming that low-return and high-return behaviors share a common high-level latent decision space—the geometric premise that makes directional quality-uplift transport feasible.

Together, these results establish three conclusions: (1) the Source-to-Positive Alignment branch constructs a non-trivial structure-forming flow from a zero-value baseline to high-value behavioral regions, (2) low-return and high-return behaviors share a common high-level latent decision space, providing the geometric premise for directional quality uplift, and (3) the dual-branch design achieves complementary roles—structure formation and directional improvement—within this shared space.

\subsection{Hyperparameter Sensitivity Analysis}

Sensitivity experiments are conducted on three medium-replay tasks (HalfCheetah, Hopper, Walker2d) to study high-level time scale and trajectory coverage ratio independently of data distribution changes.

\subsubsection{Chunk Length and High-Level Context}

Table~\ref{tab:chunk_horizon} compares $(H,K_{\max})\in\{(5,200),(10,100),(20,50)\}$ at approximately equal total primitive-step coverage. The configuration $H{=}10,K_{\max}{=}100$ achieves the best balance. Both finer ($H{=}5$) and coarser ($H{=}20$) granularities slightly reduce performance, indicating that moderate temporal abstraction is most suitable for high-level latent decision flow mapping.

\begin{table}[htbp]
\centering
\caption{Results for Different Chunk Lengths and Maximum High-Level Chunk Counts}
\label{tab:chunk_horizon}
\begin{tabular}{lrrr}
\hline
Dataset & H=5, $K_{\max}$=200 & H=10, $K_{\max}$=100 & H=20, $K_{\max}$=50 \\
\hline
HalfCheetah MR & 43.03 ± 0.35 & \textbf{43.81 ± 0.30} & 42.95 ± 0.70 \\
Hopper MR & 100.03 ± 1.39 & \textbf{104.36 ± 2.12} & 99.41 ± 2.62 \\
Walker2d MR & 91.73 ± 1.23 & \textbf{96.24 ± 1.02} & 93.28 ± 1.85 \\
\hline
\end{tabular}
\end{table}

Table~\ref{tab:max_chunks} fixes $H{=}10$ and varies $K_{\max}\in\{25,50,100\}$. Performance degrades sharply at $K_{\max}{=}25$ and $K_{\max}{=}50$, confirming that the model requires sufficient high-level context to capture cross-stage task structure. $K_{\max}{=}100$ covers the full 1000-step trajectory horizon.

\begin{table}[htbp]
\centering
\caption{Results for Different $K_{\max}$ with Fixed $H=10$}
\label{tab:max_chunks}
\begin{tabular}{lrrr}
\hline
Dataset & $K_{\max}$=25 & $K_{\max}$=50 & $K_{\max}$=100 \\
\hline
HalfCheetah MR & 11.74 ± 0.14 & 22.07 ± 0.22 & \textbf{43.81 ± 0.30} \\
Hopper MR & 25.69 ± 0.43 & 52.02 ± 0.45 & \textbf{104.36 ± 2.12} \\
Walker2d MR & 21.13 ± 0.86 & 42.58 ± 0.29 & \textbf{96.24 ± 1.02} \\
\hline
\end{tabular}
\end{table}

\subsubsection{Trajectory Coverage Ratio}

Table~\ref{tab:coverage_ratio} compares $\rho_{\mathrm{traj}}\in\{0.1,0.2,0.5,0.8\}$. Ratios of 0.1 and 0.5 both achieve performance close to the default $\rho_{\mathrm{traj}}{=}0.2$, showing that \ourmodel{} can learn stable suboptimal-to-high-value flow mapping directions from small high-value timestep budgets. However, $\rho_{\mathrm{traj}}{=}0.8$ causes sharp drops on Hopper and Walker2d, as an excessively large high-value pool reduces the directional contrast in suboptimal-to-high-value pairing.

\begin{table}[htbp]
\centering
\caption{Results for Different High-Value Timestep Coverage Ratios}
\label{tab:coverage_ratio}
\begin{tabular}{lrrrr}
\hline
Dataset & $\rho$=0.1 & $\rho$=0.2 & $\rho$=0.5 & $\rho$=0.8 \\
\hline
HalfCheetah MR & 42.32 ± 0.36 & \textbf{43.81 ± 0.30} & 43.00 ± 0.51 & 42.66 ± 0.30 \\
Hopper MR & 101.45 ± 2.33 & \textbf{104.36 ± 2.12} & 99.49 ± 3.48 & 66.24 ± 5.75 \\
Walker2d MR & 96.77 ± 2.34 & \textbf{96.24 ± 1.02} & 95.70 ± 2.17 & 85.92 ± 8.10 \\
\hline
\end{tabular}
\end{table}

\section{Conclusion}
\label{sec:conclusion}

This paper proposes \ourmodel{}, a one-step policy optimization method for offline reinforcement learning in high-level latent decision space. \ourmodel{} reformulates offline RL as a generative policy optimization problem from suboptimal to optimal, mapping trajectory chunks to high-level decision representations through a Chunk-Level Latent Decision Encoder, and learning a shared conditional mean velocity field in this space. The method completes decision refinement through a dual-branch flow mapping mechanism: the first branch transports Gaussian-Induced Decision Sources toward positive behavioral decision regions, establishing task-relevant decision structures. The second branch transports suboptimal decisions toward high-value decision regions, distilling superior behavioral patterns from existing experiences. Hierarchical chunk-level advantage estimation further guides the flow mapping toward high-return regions. Unlike multi-step denoising diffusion policies and flow matching policies requiring multi-step ODE integration, \ourmodel{} directly predicts conditional mean velocity over the unit time interval $\Delta t=1$ and completes decision refinement through one-step Euler integration, requiring neither iterative sampling nor backpropagation through time.

On 9 Gym-MuJoCo tasks from the D4RL benchmark, \ourmodel{} achieves an average score of 90.91, surpassing FlowQ (85.6) and SSCP (87.9), and significantly outperforming FQL (79.2). The high-level latent decision representation compresses sequence length from original action steps $T$ to $K \approx T/H$, significantly reducing computational overhead in long-horizon tasks. The suboptimal-to-high-value transport mechanism effectively utilizes low-quality trajectories for policy improvement, making \ourmodel{} particularly effective in long-horizon continuous control tasks. Future research can extend \ourmodel{} to goal-conditioned RL and multi-task RL scenarios, learning conditional mean velocity fields to support diverse goal-oriented behaviors and further advancing the applicability of offline RL in real-world domains. Nevertheless, the high-level latent decision representations of \ourmodel{} are constrained by the diversity of the offline dataset, and excessively narrow behavioral support may limit the quality of the learned decision space. Addressing this through meta-learning initialization, data augmentation, or VLM-grounded latent priors remains an open direction, as does extending the framework to goal-conditioned RL and multi-task RL settings to support diverse goal-oriented behaviors.\looseness=-1

\bibliographystyle{unsrtnat}
\bibliography{references}

\end{document}